\documentclass{article}
\usepackage{spconf,amsmath,amssymb,mathrsfs}
\usepackage{microtype}
\usepackage{graphicx,epsfig}
\usepackage[boxed]{algorithm2e}
\usepackage[caption=false,font=footnotesize]{subfig}

\def\x{\boldsymbol{x}}
\def\y{\boldsymbol{y}}
\def\z{\boldsymbol{z}}
\def\a{\boldsymbol{a}}
\def\r{\boldsymbol{r}}
\def\u{\boldsymbol{u}}
\def\v{\boldsymbol{v}}
\def\f{\boldsymbol{f}}
\def\g{\boldsymbol{g}}
\def\P{\mathbf{P}}
\def\C{\mathcal{C}}

\newtheorem{theorem}{Theorem}[section]
\newtheorem{definition}[theorem]{Definition}

\title{A new ADMM algorithm for the Euclidean median and its application to robust patch regression}
\name{Kunal N. Chaudhury\thanks{Correspondence: kunal@ee.iisc.ernet.in. The first author was partially supported by the Startup Grant provided by the Indian Institute of Science.}\hspace{12mm} K. R. Ramakrishnan}
\address{Department of Electrical Engineering, Indian Institute of Science, Bangalore, India}

\begin{document}
\ninept

\maketitle
\begin{abstract}
The Euclidean Median (EM) of a set of points $\Omega$ in an Euclidean space is the point $\x$ minimizing the (weighted) sum of the Euclidean distances of $\x$ to the points in $\Omega$. 
While there exits no closed-form expression for the EM, it can nevertheless be computed using iterative  methods
such as the Weiszfeld algorithm. 
The EM has classically been used as a robust estimator of centrality for multivariate data.
It was recently demonstrated that the EM can be used to perform robust patch-based denoising of images by generalizing  the popular Non-Local Means algorithm. 
In this paper, we propose a novel algorithm for computing the EM (and its box-constrained counterpart) using variable splitting and the method of augmented Lagrangian. 
The attractive feature of this approach is that  the subproblems involved in the ADMM-based optimization of the augmented Lagrangian can be resolved using simple closed-form projections. 
The proposed ADMM solver is used for robust patch-based image denoising and is shown to exhibit faster convergence compared to an existing solver.
\end{abstract}
\begin{keywords}
Image denoising, patch-based algorithm, robustness, Euclidean median, variable splitting, augmented Lagrangian, alternating direction method of multipliers (ADMM), convergence. 
\end{keywords}

\section{Introduction}
\label{sec:intro}

Suppose we are given a set of points $\a_1,\ldots,\a_n \in \mathbf{R}^d$, and some non-negative weights $w_1,\ldots,w_n$ to these points. Consider the problem
\begin{equation}
\label{EM}
\begin{aligned}
 \underset{\x \in \C}{\text{min}} \quad \sum_{k=1}^n w_k \  \lVert \x - \a_k \rVert_2 
\end{aligned}
\end{equation}
where $\lVert \cdot  \rVert_2$ is the Euclidean norm and $\C \subset \mathbf{R}^d$ is  closed and convex. 
This is a convex optimization problem.
The (global) minimizer of \eqref{EM} when $\C$ is the entire space is called the \textit{Euclidean median} (also referred to as the geometric median) \cite{Lopuhaa1991}. 
The Euclidean median has classically been used  as a robust estimator of centrality for multivariate data \cite{Lopuhaa1991}. This multivariate generalization of the scalar median also comes up in transport engineering, where \eqref{EM} is used to model the problem of locating a facility that minimizes the cost of transportation \cite{Eiselt2011}. More recently, 
it was demonstrated in \cite{Chaudhury2012,Chaudhury2013a} that the Euclidean median and its non-convex extensions can be used to perform robust patch-based regression for image denoising by generalizing the poplar Non-Local Means algorithm \cite{BCM2005,BCM2010}.

Note that the minimizer of \eqref{EM} is simply the projection of the Euclidean median onto $\C$. Unfortunately, there is no simple closed-form expression for the Euclidean median, even when it is unique \cite{Bose2003}. Nevertheless, the Euclidean median can be computed using numerical methods such as the iteratively reweighted least squares (IRLS). One such iterative method is the so-called Weiszfeld algorithm \cite{Weiszfeld1937}. This particular algorithm, however, is known to be prone to convergence problems \cite{Eiselt2011}. Geometric optimization-based algorithms have also be proposed for approximating the Euclidean median \cite{Bose2003}. More recently, motivated by the use of IRLS
for sparse signal processing and $L_1$  minimization \cite{CW2008,DDFG2009}, an IRLS-based algorithm for approximating the Euclidean median was proposed in \cite{Chaudhury2012}. More precisely, here the authors consider a smooth convex surrogate of \eqref{EM}, namely,
 \begin{equation}
 \label{EM_surrogate}
 \sum_{k=1}^n w_k \  (\lVert \x - \a_k \rVert_2^2 + \varepsilon)^{1/2} \quad \quad (\varepsilon >0),
\end{equation}
and describe an iterative method for optimizing \eqref{EM_surrogate}. The convergence properties of this iterative algorithm was later studied in \cite{Chaudhury2013}. 

 In the last few years, there has been a lot of renewed interest in the use of variable splitting, the method of multipliers, and the alternating direction method of multipliers (ADMM) \cite{Eckstein1992} for  various non-smooth optimization problems in signal processing \cite{Combettes2011,Afonso2010,Afonso2011} . Motivated by this line of work, we introduce a novel algorithm for approximating the solution of \eqref{EM} using variable splitting and the augmented Lagrangian in Section \ref{sec:ADMM}. An important technical distinction of the present approach is that, instead of working with the smooth surrogate \eqref{EM_surrogate}, we directly address the original non-smooth objective in \eqref{EM}.  
The attractive feature here is that the subproblems involved in the ADMM-based optimization of the augmented Lagrangian can be resolved using simple closed-form projections. In Section \ref{sec:results}, the proposed algorithm is used for robust patch-based denoising of images following the proposal in \cite{Chaudhury2012}. In particular, we incorporate the information that the pixels of the original (clean) patches take values in a given dynamic range (for example, in [0,255] for grayscale images) using an appropriately defined $\C$. Numerical experiments show that the iterates of the proposed algorithm converge much more rapidly than the IRLS iterates for the denoising method in question. Moreover, the reconstruction quality of the denoised image obtained after few ADMM iterations  is often substantially better than the IRLS counterpart.

\section{Euclidean Median using ADMM}
\label{sec:ADMM}

Since we will primarily work with \eqref{EM}  in this paper, we will overload terminology and refer to the minimizer of \eqref{EM} as the Euclidean median. 
The main idea behind the ADMM algorithm is to decompose the objective in \eqref{EM} into a sum of independent objectives, which can then be optimized separately. 
This is precisely achieved using variable splitting \cite{Combettes2011,Afonso2010}.
In particular, note that we can equivalently formulate \eqref{EM} as
\begin{eqnarray}
\label{splitting}
\begin{aligned}
& \underset{\z, \x_1,\ldots,\x_n}{\text{min}}
& & \iota_{\C}(\z) + \sum_{k=1}^n \ f_k(\x_k)   \\
& \text{subject to}
& & \x_k - \z =0 \quad \quad (k=1, \ldots, n),
\end{aligned}
\end{eqnarray}
where  
\begin{equation}
\label{deffk}
f_k(\x_k) = w_k   \lVert \x_k - \a_k \rVert_2,
\end{equation}
and $\iota_{\C}$ is the indicator function of $\C$ \cite{Combettes2011}. In other words, we artificially introduce one local variable for each term of the objective, and a common global variable that forces the local variables to be equal \cite{Boyd2011}. The advantage of this decomposition will be evident shortly. 

Note that in the process of decomposing the objective, we have introduced additional constraints. 
While one could potentially use any of the existing constrained optimization methods to address \eqref{splitting}, we will adopt the method of augmented Lagrangian which (among other things) is particularly tailored to the separable structure of the objective.
The augmented Lagrangian for \eqref{splitting} can be written as
\begin{equation}
 \mathcal{L} =   \iota_{\C}(\z)  + \sum_{k=1}^n \left\{ f_k(\x_k) + \y_k^{T}(\x_k - \z) + \frac{\mu}{2}  \lVert \x_k - \z \rVert^2_2 \right\}
\end{equation}
where $\y_1,\ldots,\y_n$ are the Lagrange multipliers corresponding to the equality constraints in \eqref{splitting}, and $\mu > 0$ is a penalty parameter \cite{Eckstein1992,Combettes2011}.

We next use the machinery of generalized ADMM to optimize $ \mathcal{L}$ jointly with respect to the variables and the Lagrange multipliers. The  generalized ADMM algorithm proceeds as follows \cite{Combettes2011,Boyd2011}: In each iteration, we sequentially minimize $\mathcal{L}$ with respect to the variables $\x_1,\x_2,\ldots,\x_n,\z$ in a Gauss-Seidel fashion (keeping the remaining variables fixed), and then we update the Lagrange multipliers $\y_1,\ldots,\y_n$ using a fixed dual-ascent rule. In particular, note that the minimization over $\x_k$ at the $(t+1)$-th iteration is given by
\begin{equation}
\label{updatex}
\x_k^{(t+1)} = \underset{\x}{\text{argmin}} \ \ f_k(\x) + (\x - \z^{(t)})^{T}\y_k^{(t)} + \frac{\mu}{2}  \lVert \x - \z^{(t)} \rVert^2_2
\end{equation}
where $ \z^{(t)}$ denotes the variable $\z$ at the end of the $t$-th iteration. A similar notation is used to denote the Lagrange multipliers at the end of the $t$-th iteration.

Next, by combining the quadratic and linear terms and discarding terms that do not depend on $\z$, we can express the partial minimization over $\z$  as 
\begin{eqnarray*}
\begin{aligned}
\z^{(t+1)}  &= \underset{\z}{\text{argmin}} \  \ \iota_{\C}(\z) + \frac{\mu}{2}   \sum_{k=1}^N \ \lVert \z - \x^{(t+1)}_k - \mu^{-1}\y^{(t)}_k \rVert^2_2 \\
& = \underset{\z \in \C}{\text{argmin}}  \ \  \lVert \z - \z^{(t+1)}_0 \rVert^2_2
\end{aligned}
\end{eqnarray*}
where
\begin{equation*}
 \z_0^{(t+1)} = \frac{1}{N} \sum_{k=1}^N  \left( \x_k^{(t+1)} + \mu^{-1}  \y_k^{(t)}\right).
\end{equation*}
Therefore,
\begin{equation}
\label{updatez}
\z^{(t+1)}  = \mathscr{P}_{\C}( \z_0^{(t+1)})
\end{equation}
where $\mathscr{P}_{\C}(\x)$ denotes the  projection of $\x$ onto the convex set $\C$.
Note that we have used the most recent $\x_k$ variables in \eqref{updatez}.
 Finally, we update the Lagrange multipliers using the following standard rule:
\begin{equation}
\label{updatey}
\y_k^{(t+1)} = \y_k^{(t)} + \mu \ (\x_k^{(t+1)} - \z^{(t+1)} ).
\end{equation}

We loop over \eqref{updatex}, \eqref{updatez} and \eqref{updatey} until some convergence criteria is met, or up to a fixed number of iterations. 
For a detailed exposition on the augmented Lagrangian, the method of multipliers, and the ADMM, we refer the interested reader to \cite{Eckstein1992,Combettes2011,Boyd2011}.

We now focus on the update in \eqref{updatex}. By combining the linear and quadratic terms (and discarding terms that do not depend on $\x$), we can express \eqref{updatex} in the following general form:
\begin{equation}
\label{prox}
\x^{\star} =  \ \underset{\x}{\text{argmin}} \ f(\x) + \frac{1}{2} \lVert \x - \v \rVert^2_2 \\
\end{equation}
where
\begin{equation}
\label{deff}
f(\x) = \lambda \ \lVert \x - \u \rVert_2.
\end{equation}
Note that the objective in \eqref{prox} is closed, proper, and strongly convex, and hence $\x^{\star}$ exists and is unique. In fact, we immediately recognize $\x^{\star}$ to 
be the \textit{proximal map} of $f$ evaluated at $\u$ \cite{Combettes2011,Rockafellar}.
\begin{definition} For any $f : \mathbf{R}^d \mapsto (-\infty,\infty]$ that is closed, proper, and convex, the proximal map $\Psi_f : \mathbf{R}^d \mapsto \mathbf{R}^d$ is defined to be
\begin{equation}
\label{defprox}
\Psi_f (\v) = \underset{\x}{\mathrm{argmin}} \ f(\x) + \frac{1}{2} \lVert \x - \v \rVert^2_2 \qquad (\v \in \mathbf{R}^d).
\end{equation}
\end{definition}
As per this definition, $\x^{\star} = \Psi_f(\v)$.

Note that \eqref{deff} involves  the $L_2$ norm, which unlike the $L_1$ norm is not separable. A coordinate-wise minimization of \eqref{prox} is thus not possible (as is the case for the $L_1$ norm leading to the well-known shrinkage function). However, we have the following result on proximal maps \cite{Combettes2011,Rockafellar}.
\begin{theorem} Let $f : \mathbf{R}^d \mapsto (-\infty,\infty]$ be closed, proper, and convex. Then we can decompose any $\x \in \mathbf{R}^d$ as
\begin{equation}
\label{MD}
\x = \Psi_f(\x) + \Psi_{f^{\star}}(\x),
\end{equation}
where $f^{\star} : \mathbf{R}^d \mapsto (-\infty,\infty]$ is the convex conjugate of $f$,
\begin{equation}
\label{cc}
f^{\star}(\y) =  \ \underset{\x}{\mathrm{max}} \ \ \x^{T}\y - f(\x) \quad  \quad(\y \in \mathbf{R}^d).
\end{equation}
\end{theorem}
The usefulness of \eqref{MD} is that both $f^{\star}$ and and its proximal map can be obtained in closed-form. Indeed, by change-of-variables,
\begin{eqnarray*}
\begin{aligned}
f^{\star}(\y) &=  \ \underset{\x}{\mathrm{max}} \ \ \x^{T}\y -  \lambda \ \lVert \x - \u \rVert_2 \\ 
&= \  \u^{T}\y + \ \underset{\x}{\mathrm{max}} \ \ \x^{T}\y -  \lambda \ \lVert \x  \rVert_2.
\end{aligned}
\end{eqnarray*}
Now, if $ \lVert \y  \rVert_2 \leq \lambda$, then by Cauchy-Schwarz, $\x^{T}\y \leq \lambda \ \lVert \x  \rVert_2$. In this case, the maximum over $\x$ is $0$.
On the other hand, if $ \lVert \y  \rVert_2 > \lambda$, then setting $\x = t \y \ (t>0)$, we have
\begin{equation*}
\x^{T}\y -  \lambda \ \lVert \x  \rVert_2 = t  \ \lVert \y  \rVert_2 \ ( \lVert \y  \rVert_2  - \lambda) > 0,
\end{equation*}
which can be made arbitrarily large by letting $t \rightarrow \infty$. Thus,
\begin{equation}
\label{CF}
f^{\star}(\y) = \u^{T}\y + \iota_{\mathcal{B}}(\y),
\end{equation}
where $\iota_{\mathcal{B}}$ is the indicator function of the ball $\mathcal{B}$ with centre $0$ and radius $\lambda$, 
\begin{equation*}
\iota_{\mathcal{B}}(\y) = 
\begin{cases}
0 & \text{if } \  \lVert \y  \rVert_2 \leq  \lambda, \\
\infty       & \text{else}.
\end{cases}
\end{equation*}
Having obtained \eqref{CF}, we have from \eqref{defprox},
\begin{eqnarray*}
\begin{aligned}
\Psi_{f^{\star}}(\v)&  = \underset{\x}{\text{argmin}} \ \ \u^{T}\x + \iota_{\mathcal{B}}(\x)  + \frac{1}{2} \lVert \x - \v \rVert^2_2 \\ 
&=\underset{\x \in \mathcal{B}}{\text{argmin}} \ \ \frac{1}{2} \lVert \x - (\v-\u) \rVert^2_2 ,
\end{aligned}
\end{eqnarray*}
which is precisely the projection of $\v-\u$ onto $\mathcal{B}$. This is explicitly given by
\begin{equation}
\label{proxfstar}
\Psi_{f^{\star}}(\v) = \min(\lambda, \lVert \v - \u \rVert_2 ) \ \frac{\v-\u}{\lVert \v - \u \rVert_2 }.
\end{equation}
Combining \eqref{MD} with \eqref{proxfstar}, we have 
\begin{equation}
\label{closed-form}
\Psi_f(\v) = \Psi_{\lambda}(\v, \u)  = \v - \min(\lambda, \lVert \v - \u \rVert_2 ) \ \frac{\v-\u}{\lVert \v - \u \rVert_2 }.
\end{equation}
It is reassuring to note that $ \Psi_{\lambda}(\v ; \u)$ equals $\v$ when $\lambda=0$, and equals $\u$ when $\lambda=\infty$.
In the context of the ADMM update \eqref{updatex}, note that 
\begin{equation*}
\lambda = \mu^{-1}  w_k, \ \u=\a_k, \ \text{ and } \v=\z^{(t)} - \mu^{-1}  \y_k^{(t)}.
\end{equation*}
The overall ADMM algorithm (called \texttt{EM-ADMM}) for computing the Euclidean median  is summarized in Algorithm \ref{algo}.  

\SetKwFor{Loop}{loop}{}{end}
\begin{algorithm}
\KwData{Dimension $d$, points $\a_1,\ldots,\a_n \in \mathbf{R}^d$, and $\mu>0$.}
\KwResult{$\z$.}
\textbf{Initialize}: $\z, \y_1,\ldots,\y_n \in \mathbf{R}^d$\;
\Loop{}{
$\r = 0$\;
\For{$k=1,\ldots,n$}{
$\x_k  = \Psi_{\mu^{-1} w_k}(\z - \mu^{-1}  \y_k , \a_k)$\;
$\r = \r + \x_k + \mu^{-1}   \y_k$\;
}
$\z  = \mathscr{P}_{\C} (N^{-1} \r) $\;
\For{$k=1,\ldots,n$}{
$\y_k = \y_k + \mu (\x_k - \z)$\;
}
}
\caption{Euclidean Median using ADMM (\texttt{EM-ADMM})}
\label{algo}
\end{algorithm}

We next demonstrate how \texttt{EM-ADMM} can be used for robust patch-based denoising of images. We also investigate its
convergence behavior in relation to the  IRLS algorithm in \cite{Chaudhury2012}.

\section{Application: Robust Patch Regression}
\label{sec:results}

In the last ten years or so, some very effective patch-based algorithms for image denoising have been proposed \cite{BCM2005,Kervrann2006,Aharon2008,BM3D,Milanfar2013}. 
One of the outstanding proposals in this area is the Non-Local Means (NLM) algorithm \cite{BCM2005}. Several improvements and adaptations of NLM have been proposed since its inception, some of which currently offer state-of-the-art denoising results \cite{Milanfar2013}. Recently, it was demonstrated in \cite{Chaudhury2012} that the denoising performance of NLM can be further improved by introducing the
robust Euclidean median into the NLM framework. 

We consider the denoising setup where we are given a noisy image $\g = (\g_i)_{i \in I}$ ($I$ is some linear ordering of the pixels in the image) that is derived from some clean image $\f = (\f_i)_{i \in I}$ via
\begin{equation*}
\g_i = \f_i + \sigma \ \boldsymbol{\xi}_i \quad  \quad (i \in I),
\end{equation*}
where $\boldsymbol{\xi}=(\boldsymbol{\xi}_i)_{i \in I}$ is iid $\mathcal{N}(0,1)$. The denoising problem is one of estimating the unknown $\f$ from the noisy measurement $\g$. 

 \begin{figure}[htp]
\centering
\includegraphics[width=1.0\linewidth]{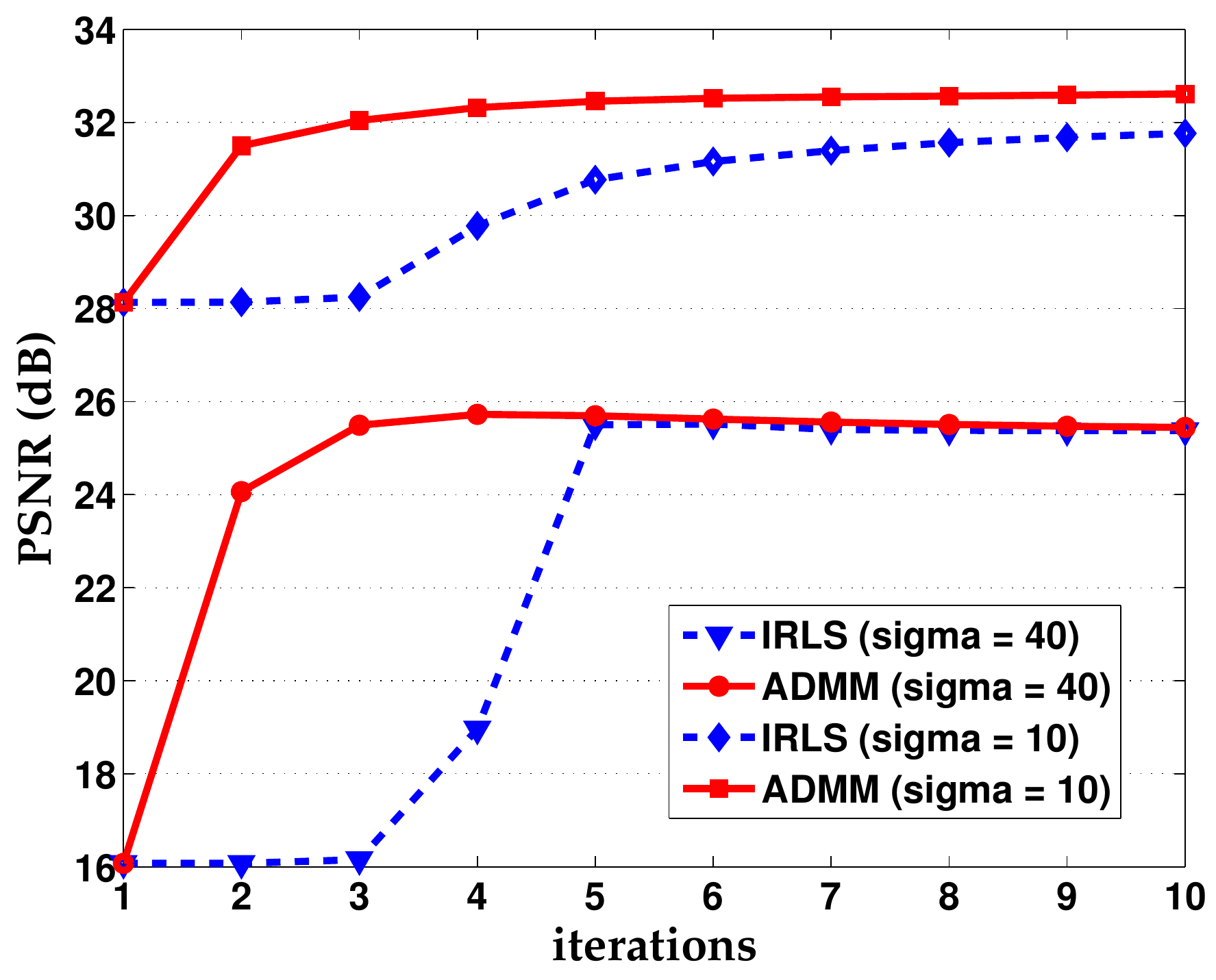}
\caption{PSNR evolution with iterations for the \textit{Barbara} image at two different noise levels. For the ADMM, we used $\mu=10^{-3}$, and for the IRLS, we set $\varepsilon=10^{-6}$.
In either case, we use the noisy patch to seed the iterations (please see main text for further details).} 
\label{psnrevolution}
\end{figure}

For any pixel $i \in I$, let patch $\P_i$ denote the restriction of $\g$ to a square window centered at $i$. Letting $k$ be the length of this window, this associates every pixel $i$ with a patch-vector $\P_i$ of length $k^2$. For $i,j \in I$, define the weights $w_{ij} = \exp(- \lVert \P_i - \P_j \rVert_2^2/h^2)$, where $h>0$ is a smoothing parameter. For every $i \in I$, we consider the following optimization problem
\begin{equation}
\label{L1regression}
\hat{\P}_i =    \underset{\P \in \mathbf{R}^{k^2} }{\mathrm{argmin}} \ \ \iota_{\C}(\P)+  \sum_{j \in S(i)} w_{ij}  \lVert \P - \P_j \rVert_2.
\end{equation}
Here $S(i)$ denotes the neighborhood  pixels of $i$, namely, those pixels that are contained in a window of size $S \times S$ centred at $i$. 
The convex set $\C$ is defined to be the collection of patches whose pixel intensities are in the dynamic range $[l,u]$; e.g., $l=0$ and $u=255$ for a grayscale image. The projection onto $\C$ can be computed coordinate-wise at negligible cost. In particular, for $1 \leq i \leq d$,
\begin{equation*}
\mathscr{P}_{\mathcal{C}}(\x)[i] =
\begin{cases}
\x[i] & \text{if } \  l \leq \x[i]  \leq u, \\
l       & \text{if } \  \x[i]  < l, \text{ and} \\
u      & \text{if } \  \x[i]  > u.
\end{cases} 
\end{equation*}

Notice that \eqref{L1regression} is exactly the optimization  in \eqref{EM}. In other words, we denoise each patch by taking the Euclidean median of its neighbouring noisy patches.  The center pixel of the denoised patch is taken to be the denoised pixel.  The overall denoising scheme is called the Non-Local Euclidean Medians (in short, NLEM) \cite{Chaudhury2012}.

\begin{table}[!htb]
\caption{Denoising performance of (a) NLM, (b) NLEM-IRLS and  (c) NLEM-ADMM at $\sigma = 10, 20, 40, 60$, and  $80$ (PSNRs averaged over $10$ noise realizations). Parameters: $S=21, k = 7, h = 10\sigma$.}  
\vspace{2mm}
\centering  
\begin{tabular}{l  c rrrrr}  

\hline 

Image & Method &\multicolumn{5}{c}{PSNR (dB)} \\

\hline

&(a)   & \bf{34.22}  &29.78        &25.20       &23.37        &22.35         \\
\textit{House} 
&(b)   &32.66    &29.01  &26.68  &24.78  &23.46   \\
&(c)   &34.13       &\bf{30.77}  &\bf{27.04}  &\bf{24.87}  &\bf{23.47}\\

\hline

&(a)   &33.24  &29.31        &26.17       &24.54        &23.64         \\
\textit{Lena} 
&(b)   &32.54    &29.48  &27.31  &25.38  &24.37   \\
&(c)   &\bf{33.57}    &\bf{30.38}  &\bf{27.59}  &\bf{25.42}  &\bf{24.38}\\

\hline

&(a)   &\bf{32.32}  &27.66        &23.11       &21.03        &19.99         \\
\textit{Peppers} 
&(b)   &30.95       &26.95  &24.31  &22.84  &21.24   \\
&(c)   &32.14       &\bf{28.54}  &\bf{25.06}  &\bf{22.98}  &\bf{21.26}\\

\hline

&(a)   &32.37  &27.39        &23.53       &22.05        &21.34         \\
\textit{Barbara} 
&(b)   &30.77       &27.28  &25.55  &23.77  &22.61   \\
&(c)   &\bf{32.43}       &\bf{28.84}  &\bf{25.67}  &\bf{23.77}  &\bf{22.62}\\

\hline

\end{tabular}
\label{tablePSNR}
\end{table}

The denoising experiments were performed on several grayscale test images. We used the standard NLM parameters for all the experiments \cite{BCM2005}: $S=21, k = 7$, and $h =10\sigma$. The proposed \texttt{EM-ADMM} algorithm is used for computing the solution of \eqref{L1regression}. 
Notice that the main computation here is that of determining the distances in \eqref{closed-form}. Since we require a total of $S^2$ such distance evaluations  in $k^2$ dimensions, the complexity is $O(S^2k^2)$ per iteration per pixel. Incidentally, this is also the complexity of the IRLS algorithm in \cite{Chaudhury2012}. For all the denoising experiments, we initialized the Lagrange multipliers  in  \texttt{EM-ADMM} to zero vectors. We considered two possible initializations for $\z$, namely, the noisy patch and the patch  obtained using NLM.  Exhaustive experiments showed that the best convergence results are obtained for $\mu \sim 10^{-3}$.
For example, figure \ref{psnrevolution} shows the evolution of the peak-signal-to-noise ratio (PSNR) over the first ten iterations obtained using the ADMM algorithm.
Also show in this figure is the evolution of the PSNR for the IRLS algorithm from \cite{Chaudhury2012}. 
In either case, we used the noisy patch as the initialization. 
Notice that the increase in PSNR is much more rapid with ADMM compared to IRLS. In fact, in about $4$ iterations, the optimal PSNR is attained with ADMM.
Notice that at $\sigma=10$, the PSNR from IRLS grows much more slowly compared to ADMM.

\begin{figure}[htp]
\centering
\includegraphics[width=1\linewidth]{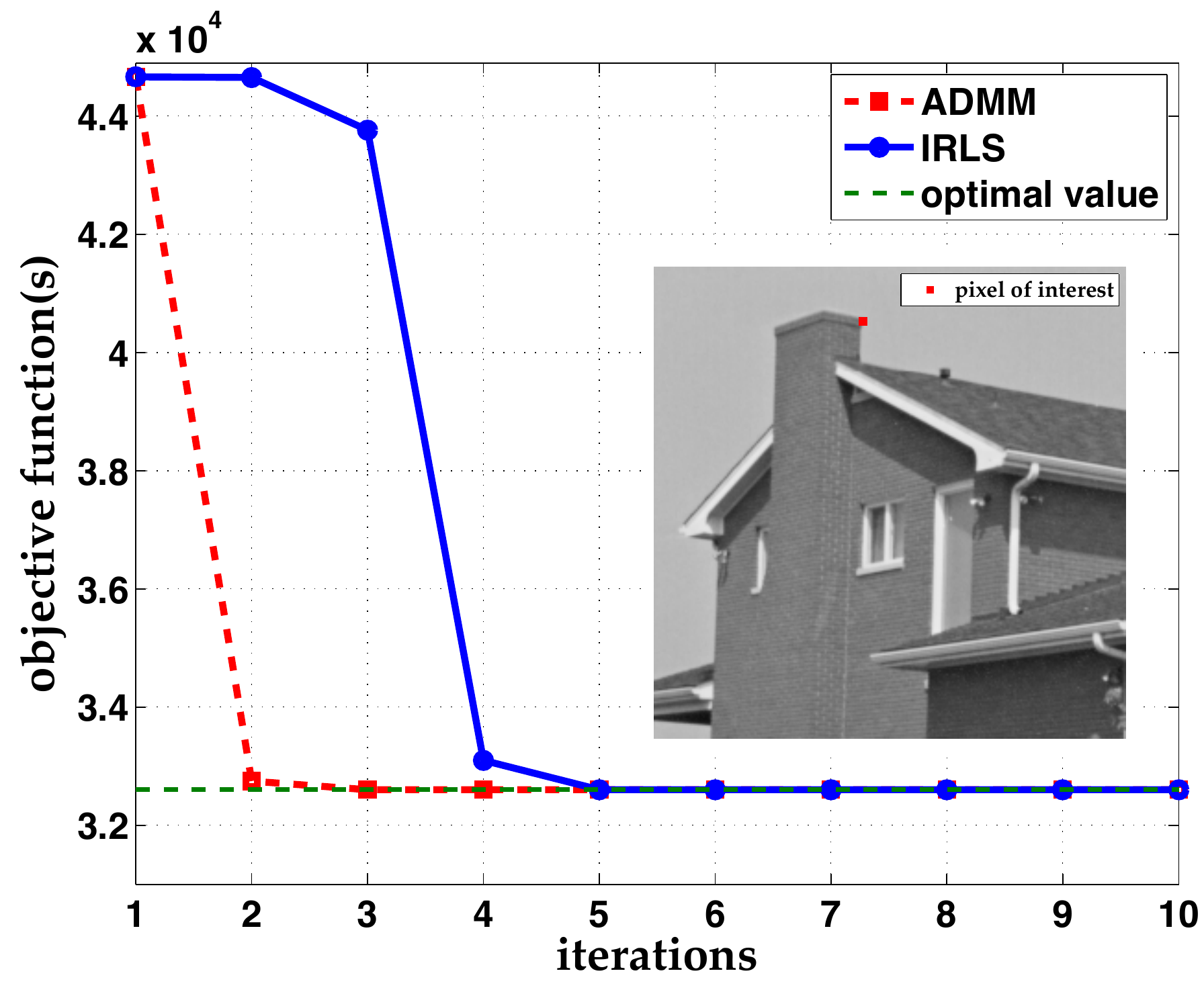}
\caption{Convergence of \texttt{EM-ADMM} and  \texttt{IRLS} at a given pixel of the \textit{House} image at $\sigma=40$. 
The inset shows the position of the particular pixel where the patch regression is performed.  
The objective functions shown in the plot correspond to \eqref{EM} and \eqref{EM_surrogate} that are  respectively optimized by ADMM and IRLS (note that the objectives are very close since we use $\varepsilon = 10^{-6}$ in  \eqref{EM_surrogate}).
The noisy image patch was used to initialize both algorithms.
For \texttt{EM-ADMM}, the objective function converges to the optimal value (up to to three decimal places) in just $2$ iterations. The convergence is relatively slow for IRLS (accuracy up to three decimals obtained after $6$ iterations).} 
\label{comparisonObj}
\end{figure}

The above observation is also supported by the convergence result shown in figure \ref{comparisonObj} for the \textit{House} image.
Notice that ADMM converges in just $2$ iterations, while IRLS takes about $6$ iterations to attain the same accuracy (we recall that the cost per iteration is comparable).
In general, we observed that the within $2$ iterations the ADMM result converges to an accuracy that is sufficient for the present denoising application. 
At higher noise levels ($\sigma>60$), about $4$ iterations are required. This applies to all the test images that we have experimented with.

\begin{figure}[htp]
\centering
\subfloat[\textit{Barbara} ($512 \times 512$).]{\includegraphics[width=0.5\linewidth]{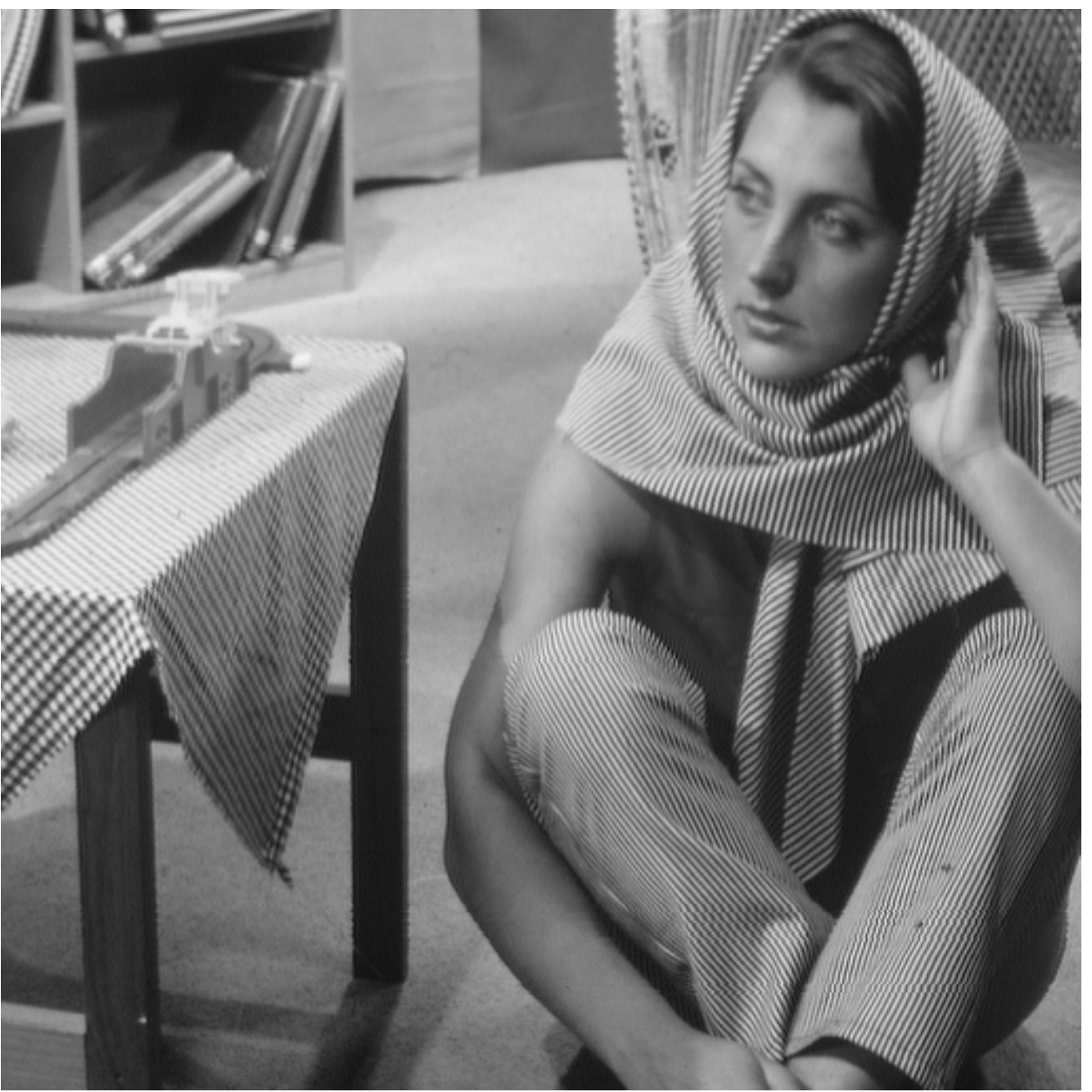}}  
\subfloat[Corrupted (PSNR = $16.15$ dB) ]{\includegraphics[width=0.5\linewidth]{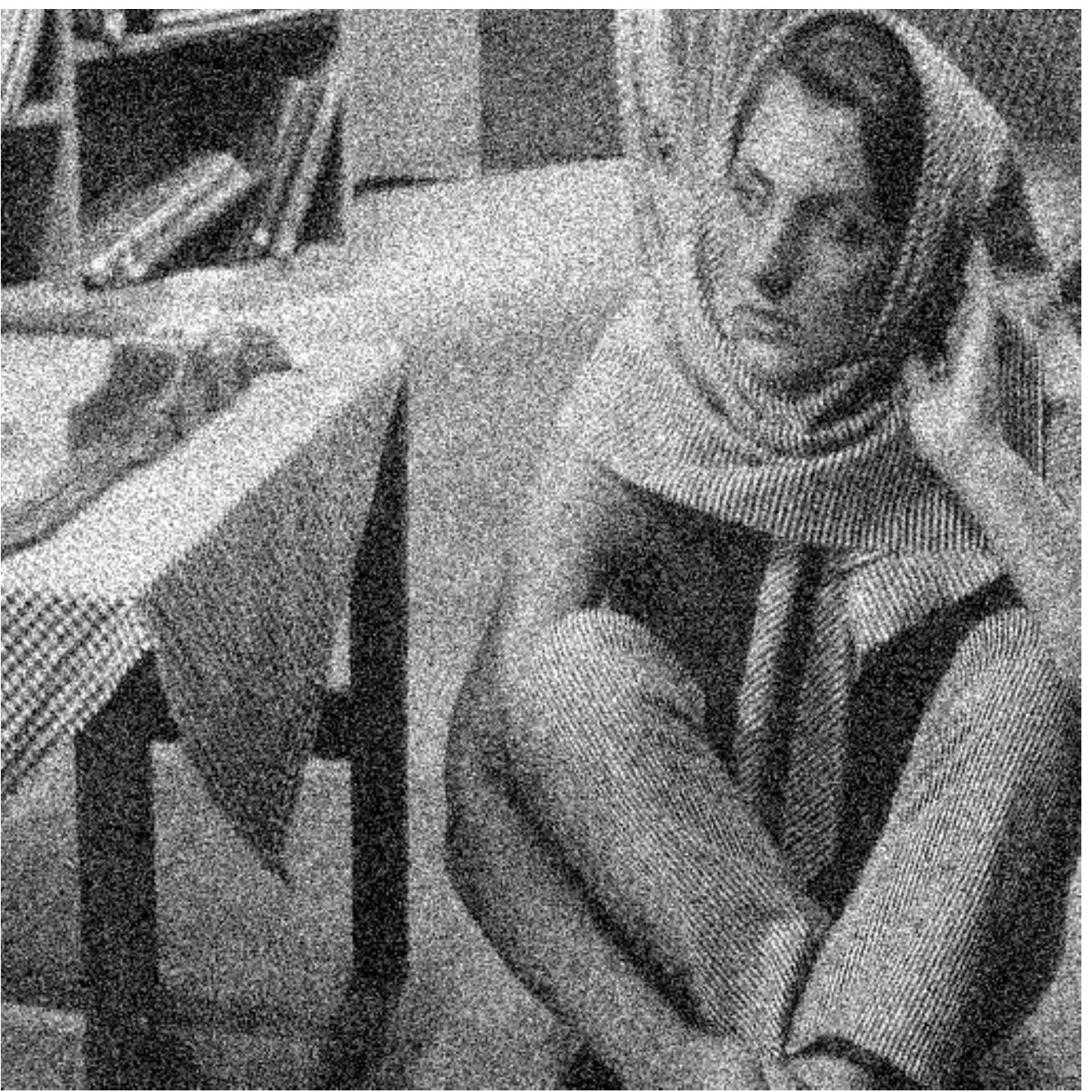}} \\ \vspace{-0.3cm} 
\subfloat[NLM (PSNR =$23.53$ dB)]{\includegraphics[width=0.5\linewidth]{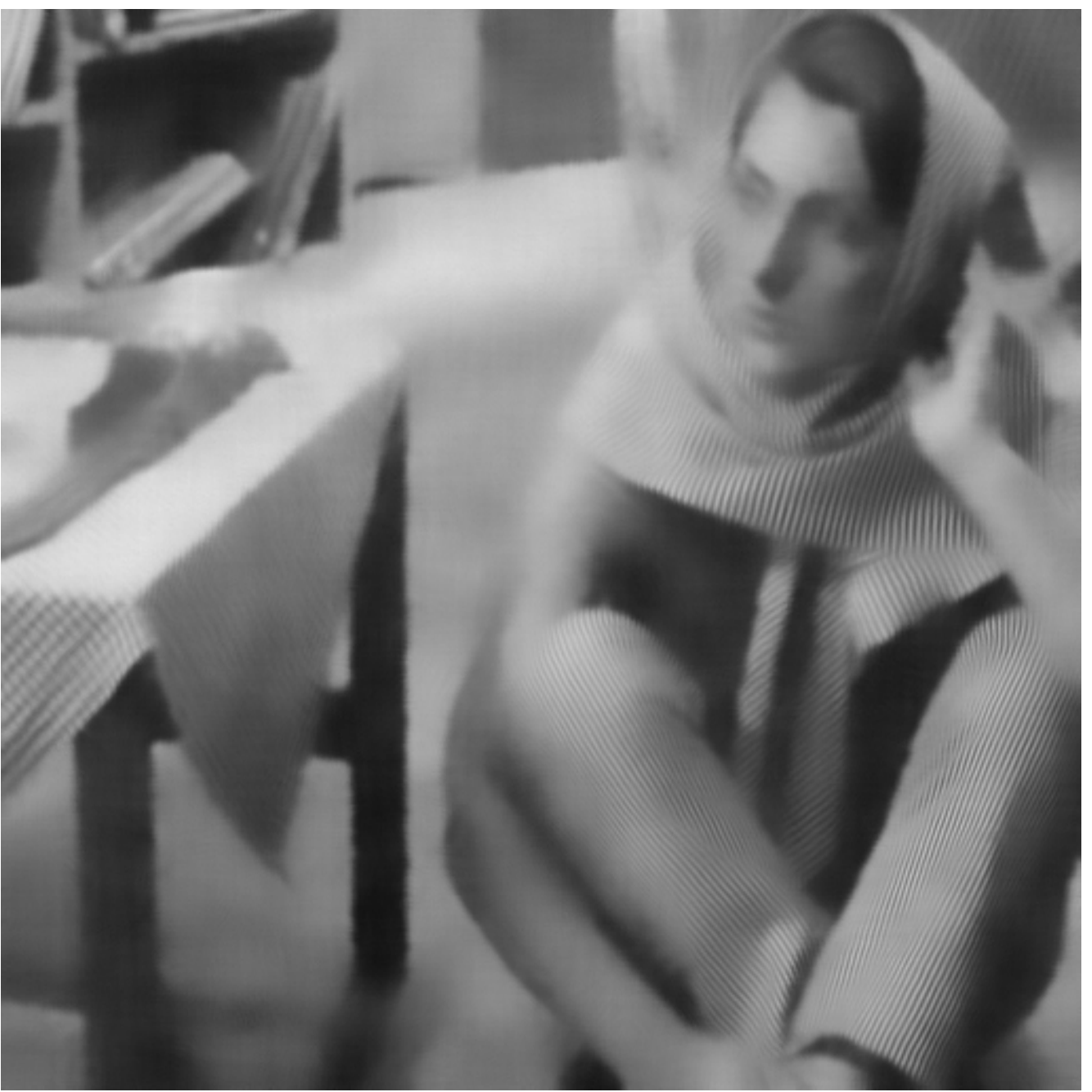}} 
\subfloat[NELM (PSNR = $\bf{25.67}$ dB).]{\includegraphics[width=0.5\linewidth]{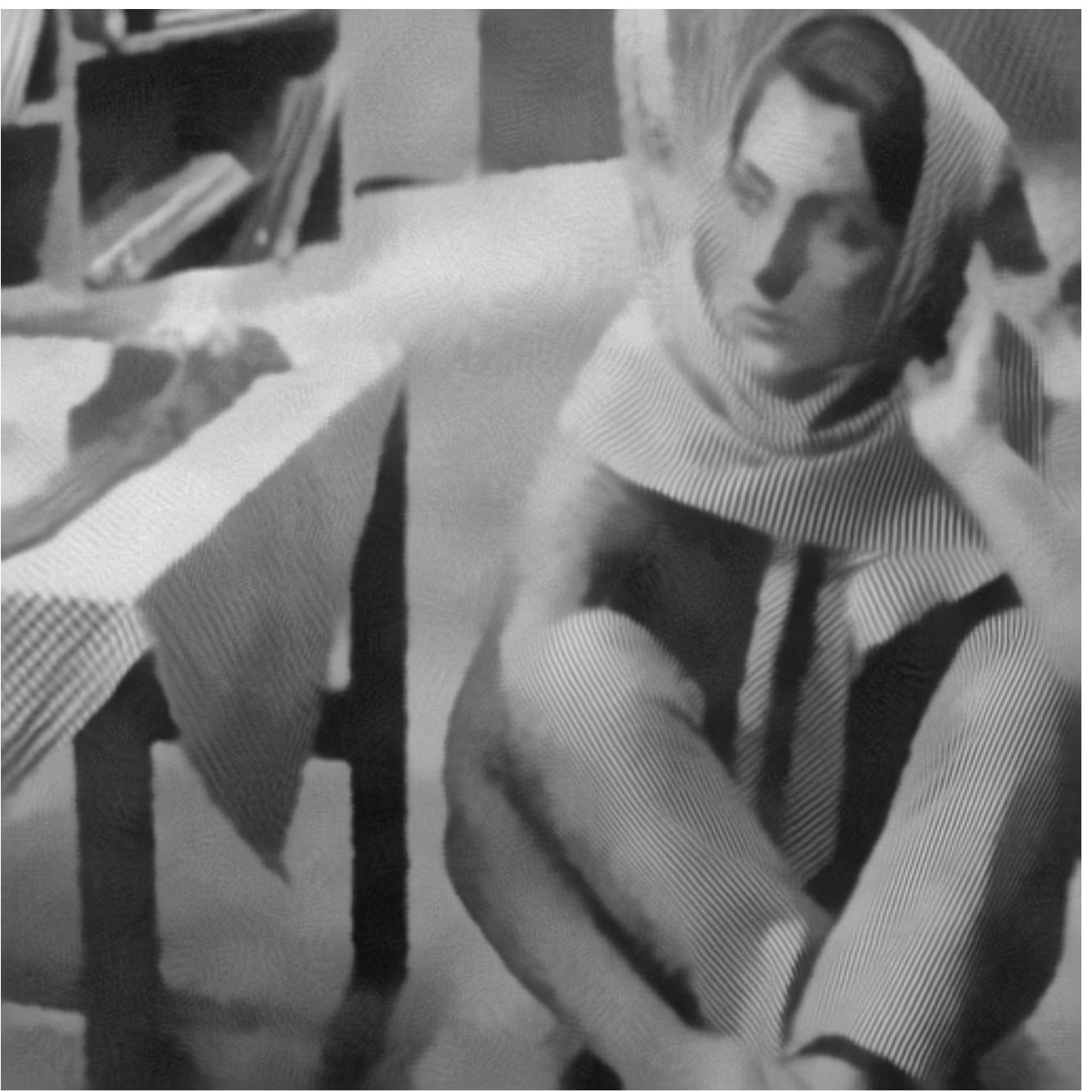}} 
\caption{Denoising results using NLM and NLEM at $\sigma=40$. We used the proposed \texttt{EM-ADMM} algorithm for computing the Euclidean medians at each pixel (please see the main text for other details about the experiment). The \texttt{EM-ADMM} algorithm was initialized using the noisy patch and run for four iterations. } 
\label{comparison1}
\end{figure}

The denoising results for some test images obtained using NLM and NLEM are provided in Table \ref{tablePSNR}. For NLEM, we used both the ADMM and IRLS solvers.
 Following the previous observations, we used $4$ iterations for both solvers.
 For $\sigma \leq 60$, we used the noisy patch to initialize the iterations, and for $\sigma > 60$ we used the NLM patch. This initialization scheme was found to give the best PSNR results.
At large noise levels, there is not much difference after $4$ iterations. However, at low noise levels, the PSNR obtained using ADMM is often substantially larger than that obtained using IRLS after $4$ iterations. The reason for this is that IRLS converges really slow in such situations. This is evident from the PSNR plots in figure \ref{psnrevolution} at $\sigma= 10$. For a visual comparison, a particular denoising result 
for the \textit{Barbara} image obtained using NLM and NLEM (using the ADMM solver) is shown in figure \ref{comparison1}. 
Roughly speaking, at the cost of just four NLMs, we are able to improve the PSNR by more than $2$ dB. Notice that the NLEM result is much more sharp compared to the NLM counterpart.

\section{Conclusion}
\label{sec:conclusion}

We proposed a new algorithm for computing the (constrained) Euclidean median using variable splitting and the augmented Lagrangian. In particular, we demonstrated how the ADMM-based optimization of the augmented Lagrangian can be resolved using simple closed-form projections. The proposed algorithm was used for image denoising using the Non-Local Euclidean Medians, and was generally found to exhibit faster convergence compared to the IRLS algorithm. One interesting direction that we did not pursue is to adapt the penalty $\mu$ at each iteration, which is known to speed up the convergence \cite{Boyd2011}. Yet another interesting direction is the use of accelerated ADMM \cite{Goldstein2012} to further speed up the convergence.

\vfill\pagebreak

\bibliographystyle{IEEEbib}

\end{document}